\documentclass[a4paper]{article}

\usepackage{INTERSPEECH2020}
\usepackage{url}

\title{Distilling the Knowledge of BERT for Sequence-to-Sequence ASR}

\name{Hayato Futami$^1$, Hirofumi Inaguma$^1$, Sei Ueno$^1$, Masato Mimura$^1$, \\ Shinsuke Sakai$^1$, Tatsuya Kawahara$^1$}
\address{
  $^1$Graduate School of Informatics, Kyoto University, Sakyo-ku, Kyoto, Japan}
\email{\{surname\}@sap.ist.i.kyoto-u.ac.jp}

\begin{document}

\maketitle
\begin{abstract}
Attention-based sequence-to-sequence (seq2seq) models have achieved promising results in automatic speech recognition (ASR).
However, as these models decode in a left-to-right way, they do not have access to context on the right.
We leverage both left and right context by applying BERT as an external language model to seq2seq ASR through knowledge distillation.
In our proposed method, BERT generates soft labels to guide the training of seq2seq ASR.
Furthermore, we leverage context beyond the current utterance as input to BERT.
Experimental evaluations show that our method significantly improves the ASR performance from the seq2seq baseline on the Corpus of Spontaneous Japanese (CSJ).
Knowledge distillation from BERT outperforms that from a transformer LM that only looks at left context.
We also show the effectiveness of leveraging context beyond the current utterance.
Our method outperforms other LM application approaches such as $n$-best rescoring and shallow fusion, while it does not require extra inference cost.
\end{abstract}

\vspace{-5pt}
\section{Introduction}
\vspace{-2pt}
End-to-end models that directly map acoustic features into symbol sequences have shown promising results in automatic speech recognition (ASR).
Compared to conventional DNN-HMM hybrid systems, end-to-end models have the advantages of a simplified architecture and fast decoding.
There are various choices when it comes to end-to-end models: connectionist temporal classification (CTC) \cite{Graves06-CTC}, attention-based sequence-to-sequence (seq2seq) models \cite{Chan16-LAS, Chorowski17-AM}, and RNN-transducer models \cite{Graves12-ST, Battenberg17-ENT}.
In this study, we adopt attention-based seq2seq models.

Seq2seq ASR models use paired speech and text for training.
In addition, unpaired text that is more readily available can be used to improve them.
An external language model (LM) is trained separately on unpaired text, and various approaches for applying the LM to ASR have been proposed.
In $n$-best rescoring, $n$-best hypotheses are obtained from ASR, followed by the addition of their LM scores, and then the best-scored hypothesis among them is selected.
Language model fusion approaches such as shallow fusion \cite{Chorowski17-TBD}, deep fusion \cite{Glehre15-DF}, and cold fusion \cite{Sriram18-CF, Toshniwal18-CT} utilize an external LM during beam-search decoding.
In shallow fusion, the linearly interpolated score from both the LM and the ASR model is used in beam search during the inference stage.
More recently, knowledge distillation \cite{Hinton15-KD} -based LM integration has been proposed \cite{Bai19-LST}.
In this approach, the LM (teacher model) provides soft labels to guide the seq2seq model (student model) training.
The LM is used during the training stage but is not required during the inference stage.

In the above-mentioned approaches that apply LM to seq2seq ASR, $n$-gram, RNNLM, or transformer \cite{Vaswani17-AIA} LM is conventionally used.
We call them ``unidirectional'' LMs, which predict each word on the basis of its left context.
In this study, we propose to apply BERT \cite{Devlin19-BERT} as an external LM.
BERT features Masked Language Modeling (MLM) in the pre-training objective, where MLM masks a word from the input and then predicts the original word.
BERT can be called a ``bidirectional'' LM that predicts each word on the basis of both its left and right context.

Seq2seq models decode in a left-to-right way, and therefore they do not have access to the right context during training or inference.
We aim to alleviate this seq2seq's left-to-right bias, by taking advantage of BERT's bidirectional nature.
$N$-best rescoring with BERT was proposed in \cite{Shin19-ESS, Salazar20-MLMS}, but the recognition result was restricted to hypotheses from left-to-right decoding.
On the other hand, BERT is difficult to use in LM fusion approaches because right (future) context that has not yet been decoded cannot be accessed during inference.
To solve these issues, we propose to apply BERT to ASR through knowledge distillation.
BERT (teacher model) provides soft labels using both left and right contexts of a current utterance for the seq2seq model (student model) training.
Furthermore, we propose to use not only right context but also context beyond utterance boundaries during the training stage.
In spontaneous ASR tasks such as presentation and conversation, the speech comprises a series of utterances.
In our proposed method, previous utterances, the current utterance, and future utterances are concatenated up to the fixed length of tokens and then fed into BERT.
BERT provides soft labels based on context that spans across utterances, which helps achieve better seq2seq ASR training.

\vspace{-5pt}
\section{Preliminaries and related work}
\vspace{-2pt}
\subsection{Sequence-to-sequence ASR}
\vspace{-2pt}
\label{sec:seq2seq-ASR}
In attention-based seq2seq ASR, we model the mapping between acoustic features and symbol sequences using two distinct networks.
One is an encoder network that transforms a sequence of acoustic features into a high-level representation.
The other is a decoder network that predicts a sequence of symbols using the encoded representation.
At each decoding step, the decoder predicts a symbol using a relevant portion of the encoded representation and previously decoded symbols.
In this study, we implemented the encoder with a multi-layer bidirectional LSTM and the decoder with a unidirectional LSTM.

Let $\bm{X} = (\bm{x}_1, ..., \bm{x}_T)$ denote a sequence of input acoustic features. Let $\bm{y} = (y_1, ..., y_N)$ denote a sequence of target symbols. The target symbols are subwords in this study, and $y_i \in \{1, ..., V\}$, where $V$ denotes the vocabulary size.
We define the seq2seq model's output probability of subword $v$ for the $i$-th target as
\begin{align}
\label{eq:P_ASR}
P_{ASR}^{(i, v)} = p(v \, | \, \bm{X}, \bm{y}_{<i})
\end{align}
$\bm{y}_{<i}$ denotes the left context of $y_i$, that is $\bm{y}_{<i} = (y_1, ..., y_{i-1})$.
During the training of a seq2seq model, we minimize the following cross-entropy objective:
\begin{align}
\label{eq:L_ASR}
\mathcal{L}_{ASR} = - \sum_{i=1}^{N} \sum_{v=1}^{V} \delta(v, y_i) \log P_{ASR}^{(i, v)}
\end{align}
where $\delta(v, y_i)$ becomes $1$ when $v = y_i$, and $0$ otherwise.

\vspace{-2pt}
\subsection{BERT}
\vspace{-2pt}
BERT \cite{Devlin19-BERT} is a mechanism for LM pre-training that consists of a multi-layer bidirectional transformer encoder \cite{Vaswani17-AIA}.
BERT can be pre-trained on a large unlabeled text and then be fine-tuned on a limited labeled text. It has shown excellent results in many downstream natural language processing tasks.
BERT's success comes from learning ``deep bidirectional'' representations.
Previous approaches to LM pre-training such as OpenAI GPT \cite{Radford18-GPT} (unidirectional) and ELMo \cite{Peters18-ELMo} (shallow concatenation of left-to-right and right-to-left RNNLMs) do not perform as well as BERT because they are not ``deeply bidirectional''.

BERT originally has two pre-training objectives: Masked Language Modeling (MLM) and Next Sentence Prediction (NSP).
MLM randomly replaces some of the input tokens with \url{[MASK]} tokens and then predicts the original word on the basis of its both left and right context.
NSP predicts whether two input sentences appear consecutively in a corpus to model sentence relationships.

\vspace{-2pt}
\subsection{Bidirectional context in seq2seq models}
\vspace{-2pt}
Seq2seq models predict each word using its left context.
Due to this autoregressive property, it is difficult for seq2seq models to leverage right context during the training and inference stages.
In seq2seq decoding, later predictions depend on the accuracy of previous predictions, and therefore the issue of error accumulation arises \cite{Bengio15-SS}.
Previous studies have addressed this issue by using right context in seq2seq ASR \cite{Mimura18-FB} and neural machine translation (NMT) \cite{Xia17-DN, Zhou19-SB}.
In \cite{Mimura18-FB}, a left-to-right and a right-to-left decoder generate $n$-best hypotheses respectively, and the two $n$-best hypotheses are then concatenated to make new hypotheses.
In \cite{Xia17-DN}, a second-pass deliberation decoder that can leverage right context was proposed. Synchronous bidirectional decoding in a single model was proposed in \cite{Zhou19-SB}.

Meanwhile, some studies have leveraged right context during the seq2seq model training by distilling the knowledge of a ``bidirectional'' teacher model \cite{Bai19-IWC, Chen20-DKB}.
In \cite{Bai19-IWC}, which is a succeeding work of \cite{Bai19-LST}, Causal clOze completeR (COR) was proposed to model both left and right context within an utterance.
In COR, the output of a stack of left-to-right transformer blocks and a stack of right-to-left ones are concatenated and fed into a subsequent fusion transformer block.
Compared to BERT, it only performs shallow concatenation of two directions of transformer blocks, and as such is not ``deeply  bidirectional''.
On the other hand, we adopt BERT, which has a simpler and more general architecture.
Furthermore, we use context that spans across utterances as input to BERT for better distillation, whereas the context is limited to the current utterance in \cite{Bai19-IWC}.
In \cite{Chen20-DKB}, a source sequence of tokens and a target sequence of tokens are fed into BERT to generate soft labels for text-to-text transduction tasks such as NMT.

\vspace{-2pt}
\subsection{Context beyond utterance boundaries in ASR}
\vspace{-2pt}
ASR is typically done at the utterance level, but context information beyond the utterance level can help improve seq2seq ASR \cite{Kim18-DCA, Masumura19-LC, Kim19-GE}.
A context vector generated from the previous utterance is incorporated into the decoder state in the current utterance in these studies.

In our method, context information beyond the utterance level is not incorporated into the ASR decoder but fed into BERT to predict better soft labels for ASR.
During inference, BERT is not used, and therefore our method does not add any extra procedure or component to utterance-level seq2seq ASR.

\begin{figure*}[t]
  \centering
\fbox{
  \includegraphics[width=0.82\linewidth]{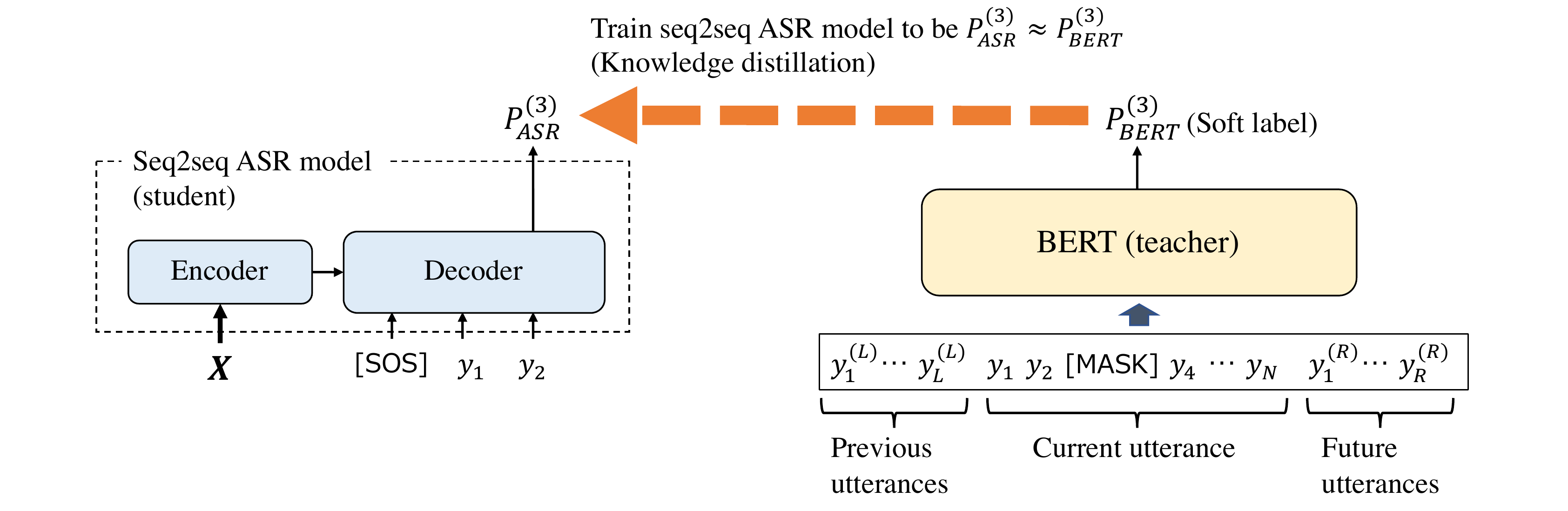}
  }
  \caption{Illustration of our proposed method. BERT generates the soft label ($=P_{BERT}^{(3)}$) using context in which $y_3$ is masked and the current utterance, previous and future utterances are concatenated.
  The target label for $P_{ASR}^{(3)}$ is given by not only the hard label ($=y_3$) but also the soft label ($=P_{BERT}^{(3)}$).}
  \label{fig:overview}
  \vspace{-10pt}
\end{figure*}

\vspace{-5pt}
\section{Proposed method}
\subsection{Pre-training BERT}
\vspace{-2pt}
In our proposed method, BERT is used as an external LM that predicts a masked word based on its context.
We need the MLM pre-training objective itself, and therefore fine-tuning for downstream tasks is not conducted.
NSP is also removed from the pre-training objective.
Following RoBERTa \cite{Liu19-RoBERTa}, BERT's input is packed with full-length sequences sampled contiguously from the corpus.

\vspace{-2pt}
\subsection{Distilling the knowledge of BERT}
\vspace{-2pt}
In our knowledge distillation, BERT serves as a teacher model and a seq2seq ASR model serves as a student model.
Pre-trained BERT provides soft labels to guide seq2seq ASR training.
These soft labels encourage the seq2seq ASR model
to generate more syntactically or semantically likely results.

Seq2seq ASR training with the knowledge of BERT is formulated as follows.
The speech in the corpus is split into a series of utterances, and the ASR model is trained on utterance-level data.
As in Section \ref{sec:seq2seq-ASR}, $\bm{X}$ denotes acoustic features in an utterance, and $\bm{y}$ denotes a label sequence corresponding to $\bm{X}$.
We utilize context beyond the current utterance as input to BERT in our method.
Let $\bm{y}^{(L)} = (y^{(L)}_1, ..., y^{(L)}_L)$ denote a subword sequence for previous (left) utterances and $\bm{y}^{(R)} = (y^{(R)}_1, ..., y^{(R)}_R)$ denote one for future (right) utterances.
The length of $\bm{y}^{(L)}$ ($=L$) and that of $\bm{y}^{(R)}$ ($=R$) are decided such that the sum of $L$, $R$, and $N$ (the label length of the current utterance) is constant (e.g. $L + R + N = 256$) and that $L$ and $R$ are the same (i.e. $L = R$).

We define BERT's output probability of subword $v$ for the $i$-th target label as
\begin{align}
\label{eq:P_BERT}
P_{BERT}^{(i, v)} &= p(v \, | \, [\bm{y}^{(L)}; \bm{y}_{\backslash i}; \bm{y}^{(R)}]) \\
\label{eq:softmax-temperature}
&= \frac{\exp{(z_v / T)}}{\sum_{j=1}^V \exp{(z_j / T)}}
\end{align}
where $z_j$ is a logit before the softmax layer and $T$ is a temperature parameter.
We obtain $\bm{y}_{\backslash i}$ by converting the $i$-th token to \url{[MASK]}, that is, $\bm{y}_{\backslash i} = (y_1, ..., y_{i-1}, $\url{[MASK]}$, y_{i+1}, ..., y_N)$.
$\bm{y}_{\backslash i}$ is concatenated with $\bm{y}^{(L)}$ and $\bm{y}^{(R)}$, then fed into BERT as $[\bm{y}^{(L)}; \bm{y}_{\backslash i}; \bm{y}^{(R)}]$.

Let $P_{ASR}^{(i)}$ and $P_{BERT}^{(i)}$ denote the probability distribution for the $i$-th target predicted by a seq2seq ASR model and by BERT, respectively.
Our goal here is to distill the knowledge of BERT and transfer it to the seq2seq ASR model by making $P_{ASR}^{(i)}$ close to $P_{BERT}^{(i)}$, as illustrated in Figure \ref{fig:overview}.
Thus, we minimize the Kullback-Leibler (KL) divergence between $P_{ASR}^{(i)}$ and $P_{BERT}^{(i)}$ for each $i$.
\begin{align}
KL(P_{BERT}^{(i)} || P_{ASR}^{(i)}) = - \sum_{v=1}^{V} P_{BERT}^{(i, v)} \log \frac{P_{ASR}^{(i, v)}}{P_{BERT}^{(i, v)}}
\end{align}

$P_{BERT}^{(i)}$ is fixed during distillation, and therefore minimizing the KL divergence over the sequence is equivalent to minimizing the following objective:
\begin{align}
\label{eq:L_KD}
\mathcal{L}_{KD} = - \sum_{i=1}^{N} \sum_{v=1}^{V} P_{BERT}^{(i, v)} \log P_{ASR}^{(i, v)}
\end{align}
The final objective is linear interpolation between $\mathcal{L}_{ASR}$ from Eq. (\ref{eq:L_ASR}) and $\mathcal{L}_{KD}$ from Eq. (\ref{eq:L_KD}).
\begin{align}
\label{eq:loss-interpolation}
\mathcal{L} = (1 - \alpha) \mathcal{L}_{ASR} + \alpha \mathcal{L}_{KD} \,\,\, (0 \leq \alpha \leq 1)
\end{align}
This can be decomposed into a soft label based on BERT $P_{BERT}^{(i, v)}$ and an one-hot label $\delta(v, y_i)$, which can be referred as a hard label.
\begin{align}
\mathcal{L} = - \sum_{i=1}^N \sum_{v=1}^V ((1 - \alpha) \delta(v, y_i) + \alpha P_{BERT}^{(i, v)}) \log P_{ASR}^{(i, v)}
\end{align}

We can pre-compute $P_{BERT}^{(i)}$ for all tokens in the training set.
For memory efficiency, we apply top-$K$ distillation \cite{Tan18-MNMT}.
We obtain the top-$K$ probabilities of BERT and normalize them for distillation.
BERT's inference is generally time-consuming because it has a large set of parameters.
However, this is not problematic in our method because we use BERT only for pre-computing soft labels of the training set and do not use it in the runtime.

\vspace{-2pt}
\subsection{Leveraging context beyond utterance boundaries}
\vspace{-2pt}
\label{sec:leverage-context-beyond}
In our method, BERT predicts soft labels on the basis of context that spans across utterances.
Tokens from previous utterances and tokens from future utterances are added to the current utterance to make up a sequence of a fixed length.
We expect two benefits from looking at context beyond utterance boundaries.
The first is that the ASR model can be trained with more informative soft labels.
It is sometimes difficult to predict words just looking at their context within the current utterance, especially in short utterances.
In this case, the top-$K$ entries in BERT's prediction get less syntactically or semantically relevant to corresponding hard labels.
Such soft labels can have an adverse effect on seq2seq ASR training.
With context beyond the current utterance, the quality of soft labels does not depend on whether the current utterance is short or long.

The other possible benefit is that we can solve the mismatch between BERT's pre-training and distillation.
While BERT is pre-trained on ``full-length'' sequences, the utterances are of various lengths.
By adding tokens from adjacent utterances up to ``full-length'', BERT is expected to perform better during distillation.

\vspace{-5pt}
\section{Experimental evaluations}
\subsection{Experimental conditions}
\vspace{-2pt}
We evaluated our method using the Corpus of Spontaneous Japanese (CSJ) \cite{maekawa03-CSJ} and the Balanced Corpus of Contemporary Written Japanese (BCCWJ) \cite{Maekawa14-BCCWJ}.
CSJ includes two subcorpora, CSJ-APS and CSJ-SPS.
CSJ-APS consists of about 240 hours of oral presentation speeches from academic meetings, and CSJ-SPS consists of about 280 hours of simulated presentation speeches on general topics.
CSJ-eval1, which is an official test set of CSJ-APS, was used for evaluation.
We also used BCCWJ-PB and BCCWJ-LB in BCCWJ as additional text for training LMs.
BCCWJ-PB consists of samples extracted from published books, and BCCWJ-LB consists of samples from books registered in libraries.
The text is tokenized using Byte Pair Encoding \cite{sennrich16-BPE} of vocabulary size 7520.
BCCWJ-PB and BCCWJ-LB have about 37M and 40M subword tokens, respectively.
The transcriptions of CSJ-APS and CSJ-SPS have about 3.9M and 4.1M subword tokens, respectively.

In our seq2seq ASR, the encoder consists of 5 layers of bidirectional LSTMs with 320 hidden states, and the decoder consists of a single LSTM layer with 320 hidden states.
We trained the seq2seq model on CSJ-APS with a batch size of 25 utterances.
The average token length of utterances was about 24 (maximum: 118, minimum: 1).
We used Adam \cite{Kingma15-Adam} with the learning rate of 1e-4 for optimizing the ASR model.
SpecAugument \cite{Park19-SA} was applied to the acoustic features.
We also applied label smoothing \cite{Szegedy16-LS}.
In target labels, the probability of 0.1 was distributed uniformly over all classes.
In decoding, we used beam search with a beam width of 5.

We trained BERT and a unidirectional transformer LM for comparison.
BERT and the transformer LM have 6 layers of transformer blocks with 512 hidden states and 8 attention heads.
We trained them on BCCWJ-PB and BCCWJ-LB first, then on the transcriptions of CSJ-APS and CSJ-SPS.
We sampled 150 sequences of length 256 for each pre-training step.
In BERT, we randomly selected 8\% of the tokens in each sequence and replaced them with \url{[MASK]} tokens.
We used Adam with the learning rate of 1e-4 with learning rate warmup over the first 10\% of total steps and linear decay.
$K$ for top-K distillation was set to 8 in all our experiments.
The temperature parameter $T$ in Eq. (\ref{eq:softmax-temperature}) and the distillation weight $\alpha$ in Eq. (\ref{eq:loss-interpolation}) were adjusted using the development set.
Our code for the proposed method is available \footnote{https://github.com/hfutami/distill-bert-for-seq2seq-asr}.

\subsection{Experimental results}
We evaluated our method through ASR experiments.
First, we compared the performances of the ASR models trained using BERT and the unidirectional transformer LM (TrfLM(uni)) as a teacher model.
We also evaluated the effectiveness of using context beyond the current utterance.
The ASR results are shown in Table \ref{tab:main}.
The result denoted as ``utterance'' in the ``Context size'' column corresponds to the ASR model guided by soft labels based on context within the current utterance.
The result denoted as ``256'' in the ``Context size'' column corresponds to that guided by soft labels based on context of length 256 that spans across utterances.
In TrfLM(uni), we added only previous utterances to the current utterance as context.
The first line in the table denotes the baseline ASR without distillation.
As shown in Table \ref{tab:main}, knowledge distillation-based LM integration consistently improved the performance of the ASR model.
We found that distillation from BERT outperformed that from the TrfLM(uni), which indicates the effectiveness of leveraging both left and right context.
We also found that incorporating context beyond the current utterance was important for distillation from BERT by comparing line 4 and 5 in the table.
This result improved the WER by $10.86\%$ relatively over the baseline.

\begin{table}[t]
  \caption{The performance for ASR trained on CSJ-APS (240h) with knowledge distillation-based LM integration. ``TrfLM(uni)'' in the ``LM'' column denotes the transformer LM.}
  \label{tab:main}
  \centering
  \begin{tabular}{ccc}
    \toprule
    {LM} & {Context size} & {WER(\%)} \\
    \midrule
    --- & --- & $10.31$ \\
    TrfLM(uni) & utterance & $9.89$ \\
    TrfLM(uni) & 256 & $10.01$ \\
    BERT & utterance & $9.53$ \\
    BERT & 256 & $\bm{9.19}$ \\
    \bottomrule
 \end{tabular}
\end{table}

\begin{table}[t]
  \caption{Ablation studies on the length of BERT's input during pre-training and distillation.}
  \label{tab:ablation}
  \centering
  \begin{tabular}{ccc}
    \toprule
    \multicolumn{2}{c}{Context size} \\ Pre-training & Distillation & WER(\%) \\
    \midrule
    64 & utterance & $9.91$ \\
    64 & 64 & $9.69$ \\ \hline
    128 & utterance & $9.62$ \\
    128 & 128 & $9.40$ \\ \hline
    256 & utterance & $9.53$ \\
    256 & 64 & $9.28$ \\
    256 & 128 & $9.28$ \\
    256 & 256 & $\bm{9.19}$ \\
    \bottomrule
 \end{tabular}
\end{table}

\begin{figure}[t]
  \centering
  \includegraphics[width=\linewidth]{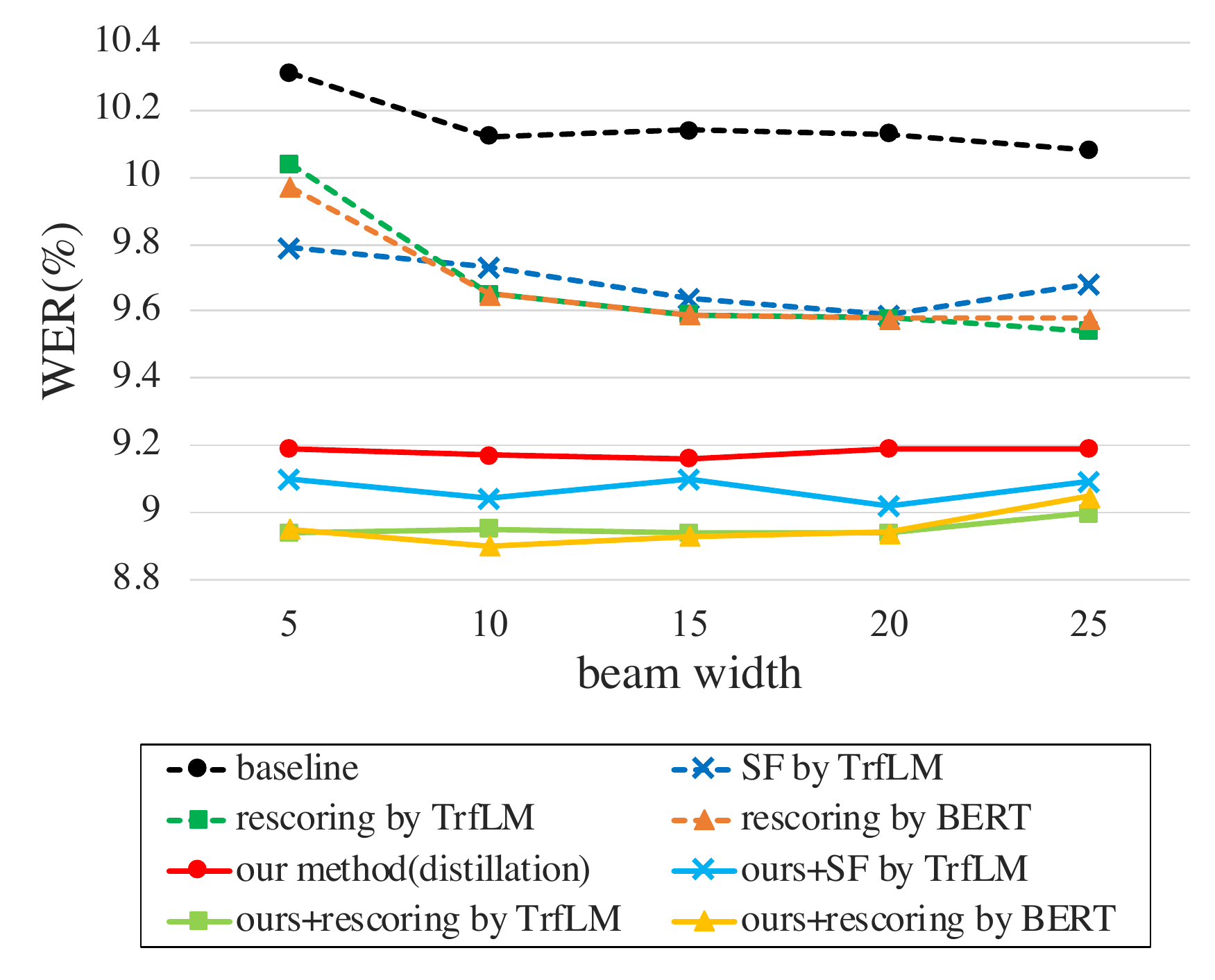}
  \caption{Comparisons and combinations with other LM application approaches. ``SF'' denotes shallow fusion.}
  \label{fig:lm-apply}
\end{figure}

\begin{table}[t]
  \caption{The performance for ASR trained on an increased amount of data (520h, both CSJ-APS and CSJ-SPS).}
  \label{tab:aps-sps}
  \centering
  \begin{tabular}{ccc}
    \toprule
    LM & Context size & WER(\%) \\
    \midrule
    --- & --- & $8.43$ \\
    BERT & 256 & $\bm{7.85}$ \\
    \bottomrule
 \end{tabular}
\end{table}

Next, we compared our method with two other LM application approaches.
Shallow fusion (SF) and $n$-best rescoring were applied to the baseline and were compared to the ASR model trained with our method (the last line in Table \ref{tab:main}).
As shown in Figure \ref{fig:lm-apply}, our method outperformed both shallow fusion and $n$-best rescoring regardless of the beam width.
We also applied shallow fusion and $n$-best rescoring to the ASR model trained through our method and obtained some improvements, which were not as large as those applied to the baseline.
This can be interpreted as the ASR model with our method had already learned the effect of applying an external LM through distillation.

Next, we conducted ablation studies on context size during pre-training and distillation. The results are shown in Table \ref{tab:ablation}.
We found that the use of longer context in the pre-training led to better ASR performance.
We also found that distillation from BERT using the same context size as pre-training performed best.

Finally, to see the effect of an increased amount of training data for ASR in our method, we trained another ASR model on both CSJ-APS and CSJ-SPS (total 520h) and evaluated the performance.
As shown in Table \ref{tab:aps-sps}, our method was still effective for this better baseline ASR model trained on an increased amount of paired data.

\section{Conclusions}
BERT can be pre-trained on a large unpaired text, and can also leverage not only left context but also right context that seq2seq ASR models do not have access to.
In this study, we have proposed a method in which the knowledge of BERT is transferred to seq2seq ASR through a knowledge distillation framework and demonstrated its effectiveness through experiments.
We found that distillation from BERT yields better ASR performance than that from the transformer LM.
We also found that the knowledge of BERT based on context that spans across utterances further improved the performance of seq2seq ASR.
Our proposed method outperformed other LM application approaches such as $n$-best rescoring and shallow fusion, including rescoring with BERT, even though our method does not require extra inference cost.
As a future work, we will investigate applying other LM pre-training mechanisms such as XLNet \cite{Yang19-XLN} and ELECTRA \cite{Clark20-ELECTRA} to ASR.

\bibliographystyle{IEEEtran}
\bibliography{mybib}

\end{document}